\documentclass[letterpaper]{article} 
\usepackage{aaai25}  
\usepackage{times}  
\usepackage{helvet}  
\usepackage{courier}  
\usepackage[hyphens]{url}  
\usepackage{graphicx} 
\usepackage{natbib}  
\usepackage{caption} 
\usepackage{algorithm}
\usepackage{algorithmic}

\usepackage{newfloat}
\usepackage{listings}
\DeclareCaptionStyle{ruled}{labelfont=normalfont,labelsep=colon,strut=off} 
\floatstyle{ruled}
\newfloat{listing}{tb}{lst}{}
\floatname{listing}{Listing}
\usepackage{amsmath}
\usepackage{amssymb}
\usepackage{booktabs}
\usepackage{multirow}

\title{Gaze Label Alignment: Alleviating Domain Shift for Gaze Estimation}

\author{Guanzhong Zeng\equalcontrib, Jingjing Wang\equalcontrib, Zefu Xu, Pengwei Yin, Wenqi Ren\thanks{Corresponding author.}, Di Xie, Jiang Zhu}
\affiliations{
    Hikvision Research Institute, Hangzhou, China \\
  \{zengguanzhong, wangjingjing9, xuzefu, yinpengwei, renwenqi, xiedi, zhujiang.hri\}@hikvision.com
}

\begin{document}

\maketitle

\begin{abstract}
Gaze estimation methods encounter significant performance deterioration when being evaluated across different domains, because of the domain gap between the testing and training data. Existing methods try to solve this issue by reducing the deviation of data distribution, however, they ignore the existence of label deviation in the data due to the acquisition mechanism of the gaze label and the individual physiological differences. In this paper, we first point out that the influence brought by the label deviation cannot be ignored, and propose a gaze label alignment algorithm (GLA) to eliminate the label distribution deviation. Specifically, we first train the feature extractor on all domains to get domain invariant features, and then select an anchor domain to train the gaze regressor. We predict the gaze label on remaining domains and use a mapping function to align the labels. Finally, these aligned labels can be used to train gaze estimation models. Therefore, our method can be combined with any existing method. Experimental results show that our GLA method can effectively alleviate the label distribution shift, and SOTA gaze estimation methods can be further improved obviously.
\end{abstract}

\section{Introduction}

\label{sec:intro}

The human gaze provides a wealth of information,
highly accurate gaze estimation can strongly support many applications, such as human-computer interaction \cite{admoni2017social, liu2021anisotropic}, augmented reality \cite{murphy2010head}, driver monitoring systems \cite{vicente2015driver,shah2022driver} and saliency prediction \cite{chang2019salgaze,sugano2012appearance}.
Recently, appearance-based gaze estimation methods \cite{liu2019differential,Cheng_2018_ECCV,crga} have achieved remarkable results with the development of deep learning. However, these methods obtain promising performance in within-domain evaluations but suffer from dramatic degradation in cross-domain evaluations due to the domain gap \cite{clipgaze}.

\begin{figure}[htp]
        \centering
        \includegraphics[width=0.35\textwidth]{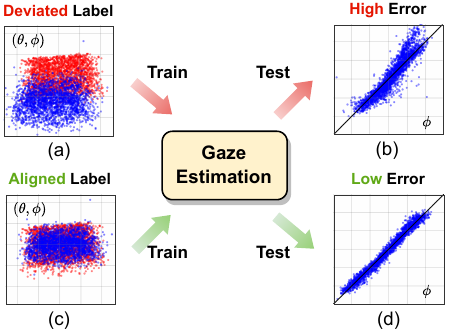}
        \vspace{-0.2cm}
        \caption{
        The red scatters in (a) and (c) mean the labels of source domain $A$ and the corresponding blue scatters are the labels of source domain $B$.
        (b) and (d) are the scatter plots of estimated label (X-axis) and ground truth label (Y-axis).
        }
        \label{fig:motivation}
        \vspace{-0.5cm}
    \end{figure}

To improve the performance in cross-domain scenarios, previous methods leverage domain adaptation approaches to reduce the domain gap. However, these methods \cite{dagen,pnpga,ruda,crga}  require target domain samples for jointly optimizing the network with source domain data, which limits its application in real world, since the source data is usually not accessible due to data privacy, and  it is not feasible to collect large amount of target domain data. To eliminate the dependency on target domain data during training, recently, some domain generalization approaches have been proposed. These methods \cite{puregaze,crga,gazeconsistence,clipgaze} do not need the target domain data, and learn domain invariant features by removing the gaze-irrelevant domain features.

Previous methods \cite{li2018deep,li2020domain,motiian2017unified,wang2022generalizing} only align the data distribution among domains and neglect the label distribution deviation. As it is infeasible to label the gaze direction manually, currently most methods get the gaze directions by building a gaze label acquisition system. The different internal and external parameters of different label acquisition systems lead to label shifts among different datasets. Moreover, even in the same label acquisition system, among different individuals, there also exists individual deviation due to the differences in eye shapes and inner eye structures. The deviation of the label causes the learned regressor to be divergent and have a relatively high error as shown in Fig.~\ref{fig:motivation}. In this paper, we point out the label distribution shift is also one of the main reasons causing performance degradation which has not been found by previous researchers in gaze estimation.

To align the label distributions, we propose a gaze label alignment algorithm. Specifically, we firstly train the feature extractor on all domains to get domain invariant features. Secondly, we select one of the domains as the anchor domain, and train the gaze regressor on the anchor domain. Thirdly, we predict the gaze label on remaining domains. then use the predicted labels and the ground truth labels on each domain to learn a mapping function to align their labels with the anchor domain. These two steps can be repeated several times. Finally, the data of all domains and aligned labels
 can be used to train the final gaze estimation model.

The main contributions can be summarized as follows:
\begin{itemize}
    \item We are the first to point out that the label distribution shift also affects the domain generalization error, and should be aligned for better performance.
    \item We propose a gaze label alignment method (GLA) to align the label distribution shift among domains, which can be combined with any gaze estimation method.
    \item Experimental results demonstrate that our GLA method can effectively alleviate the label distribution shift. By aligning the label distribution, all the SOTA gaze estimation methods can be further improved obviously.
\end{itemize}

\section{Related Works}

\subsection{Appearance-based Gaze Estimation}
Appearance-based gaze estimation models directly regress gaze direction from face images.
Since gaze labels cannot be manually annotated and data collection is expensive, in order to improve the performance of gaze estimation, early works prefer to learn more robust gaze features by relying on human prior knowledge. Zhang et al. \cite{gazenet} first proposed regress gaze direction from eye images with a deep learning network. Krafka et al. \cite{gazecapture} additionally added face image as model input to supplement more gaze-related features.  Liu et al. \cite{multitask} constructed eye landmarks regression branch to facilitate the extraction of gaze features.
Furthermore, Cheng et al. \cite{coarsetofine} designed a coarse-to-fine strategy to refine the gaze features, and Biswas et al. \cite{attention} assigned different attention weights to left and right eye features.
Although these methods have achieved good within-domain performance on limited training data, they have high generalization errors on cross-domains \cite{gazeconsistence}.

\subsection{Gaze Generalization and Adaptation}
Due to the complex changes in head pose, illumination and face appearance in the real world.
There are obvious domain gap between source domain and target domain.
Previous methods focused on eliminating the data distribution shift and obtaining domain-invariant gaze features.
For example, Cheng et al. \cite{puregaze} proposed a self-adversarial framework to purify gaze-related features and remove gaze-irrelevant features.
Xu et al. \cite{gazeconsistence} generated noisy data through adversarial attacks, endowing the model generalization ability.
Yin et al. \cite{yin2024nerf} utilized NeRF to generate diverse data to improve the performance in target domains.
In addition, Wang et al. \cite{crga} adapted to the target domain distribution by constraining the feature relationship among samples.
Liu et al. \cite{pnpga} and Cai et al. \cite{unrega} employed multiple models to reduce the uncertainty of model prediction.
Guo et al. \cite{dagen} and Bao et al. \cite{ruda} ensured the consistency of feature embedding to adapt to the target domain.
Recently, Yin et al. \cite{clipgaze, yin2025lg} leveraged pre-trained vision-language models to introduce rich general knowledge and handle diverse target domains.
However, they overlook label distribution shift between gaze domains, which also weakens the generalization capabilities of gaze estimation models.
Therefore, we analyze the root causes and impacts of label domain shift and propose a label alignment method to reduce the difference in label distribution.

\section{Method}

\subsection{Preliminaries}
A gaze estimation network consists of two parts: a feature extractor $F$ which outputs the gaze features $X=F(I)$ based on input image $I$, and a gaze regressor $W$ which estimates the gaze direction $Y=W(X)$ based on input features $X$.
Nevertheless, even the gaze estimation model achieves competitive performance in within-domain evaluation, the test error still has a high inter person variance in cross-domain evaluation \cite{zhang2018efficient, shrivastava2017learning}.
This is caused by many factors including dependencies on head poses, large eye shape variabilities, and only very subtle eye appearance changes when looking at targets separated by such small gaze angle differences \cite{liu2019differential}.
Thus, previous domain generalization methods try to find a robust feature extractor $F$, which can extract invariant features against data distribution shift among domains.
However, they neglect there is label deviation in gaze estimation due to the gaze label acquisition systems and individual physiological differences in the eyeball.
This may affect the learning of the gaze regressor and hamper the final generalization performance.

More formally, let $\mathcal{X}$ be the feature space and $\mathcal{Y}$ the label space, a domain is defined as a joint distribution $P(X,Y)$ on $\mathcal{X} \times \mathcal{Y}$. Then $P(X,Y)$ can be decomposed into $P(X,Y)=P(Y|X)P(X)$, where $P(X)$ is the marginal distribution on $X$, and $P(Y|X)$ is the posterior distribution of $Y$ given $X$. Previous domain generalization methods try to find a transformation of the data to extract feature $X$, which minimizes the difference between marginal distributions $P(X)$ of domains, while neglect the difference between $P(Y|X)$ which is caused by the gaze label acquisition systems and individual physiological differences. We call $P(X)$ the data distribution, and $P(Y|X)$ the label distribution.

\begin{figure}[htp]
  \centering
  \includegraphics[width=0.45\textwidth]{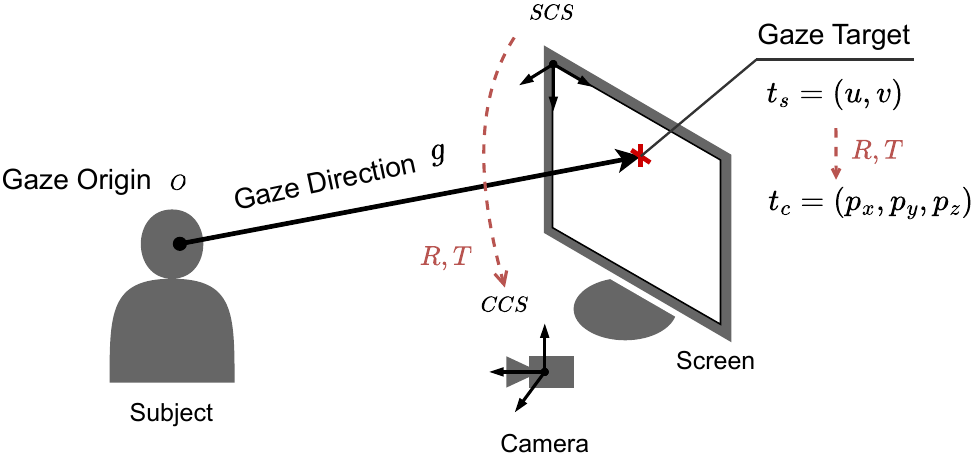}
  \vspace{-0.2cm}
  \caption{The schematic diagram of gaze direction. Gaze direction is originated from the gaze origin $\boldsymbol{o}$ and intersects with the screen at gaze target $\boldsymbol{t}_c$.}
  \label{fig:gaze}
  \vspace{-0.5cm}
\end{figure}

\subsection{What causes the Gaze Label Deviation?}
\label{sec:gaze_deviation}

\subsubsection{Individual Physiological Differences}
The main reason for the gaze label deviation between different peoples is that the visual axis is not aligned with the optical axis (related to the observed iris) \cite{pupilcenter}.
And such alignment differences are person-specific, with a standard deviation of 2 to 3 degrees among the population without eye problems \cite{liu2019differential}.
Said differently, in theory, images of two eyes with the same appearance but with different internal eyeball structure can correspond to different gaze directions, which will result in prediction uncertainties of gaze estimation model.

\subsubsection{Measurement Deviation of Data Acquisition System}
In addition, since it is infeasible to label the gaze direction manually, currently most methods get the gaze directions by building a gaze label acquisition system.
The measurement deviations of these label acquisition systems lead to label deviations between multiple gaze datasets.

Usually, the system consists of a screen and a camera, which is shown in Fig. \ref{fig:gaze}.
The screen is used to show a gaze target $\boldsymbol{t}_s=(u,v)$ in the screen coordinate system (SCS), and a user is required to look at this gaze target.
Meanwhile, a face image is captured by the camera.
The gaze direction $\boldsymbol{g}$ can be calculated as follows, and treated as the ground truth gaze label in this image.

First, the gaze target $\boldsymbol{t}_s=(u,v)$ is transformed from the SCS to $\boldsymbol{t}_c(x_t, y_t, z_t)$ in the camera coordinate system (CCS) through the extrinsic parameters between the screen and camera, i.e. $\boldsymbol{t}_c=\boldsymbol{R}[u, v, 0]^T+\boldsymbol{T}$, where the additional 0 is the $z$-axis coordinate of $\boldsymbol{t}_s$ in SCS. $\boldsymbol{R}\in \mathbb{R}^{3 \times 3}$ is the rotation matrix and $\boldsymbol{T}\in \mathbb{R}^{3 \times 1}$ is the translation matrix.
They can be derived by geometric calibration, and are usually calibrated once and applied to all gaze targets.
The gaze direction $\boldsymbol{g}$ can be defined by the 3D gaze origin $\boldsymbol{o}=(x_0, y_0, x_0)$ and the 3D gaze target $\boldsymbol{t}_c$. The 3D gaze origin $\boldsymbol{o}$ is usually defined as the face center or the eye center in CCS, and can be measured by 6-D pose estimation methods.
Given $\boldsymbol{o}$ and $\boldsymbol{t}_c$, $\boldsymbol{g}$ can be calculated as:
\begin{equation}
\boldsymbol{g} = \frac{\boldsymbol{t}_c-\boldsymbol{o}}{\left \| \boldsymbol{t}_c-\boldsymbol{o} \right \| }
\nonumber
\end{equation}

\begin{figure}[htp]
  \centering
  \includegraphics[width=0.47\textwidth]{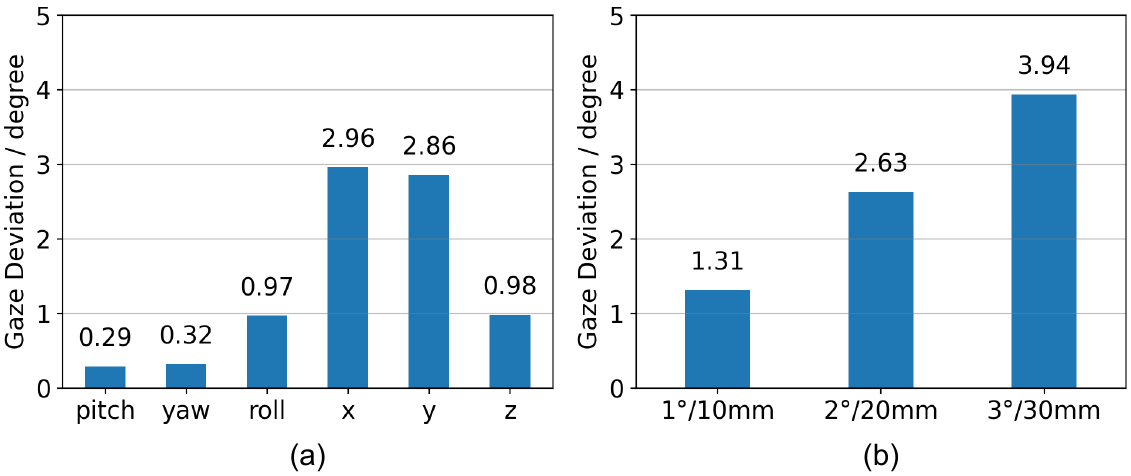}
  \vspace{-0.7cm}
  \caption{Simulation analysis of the impact of gaze label deviation. (a) is the gaze deviation caused by each single variable of calibration error. (b) is the gaze deviation of coupling multiple calibration error variables.}
  \label{fig:analysis}
  \vspace{-0.2cm}
\end{figure}

\begin{figure}[ht]
  \centering
  \includegraphics[width=0.38\textwidth]{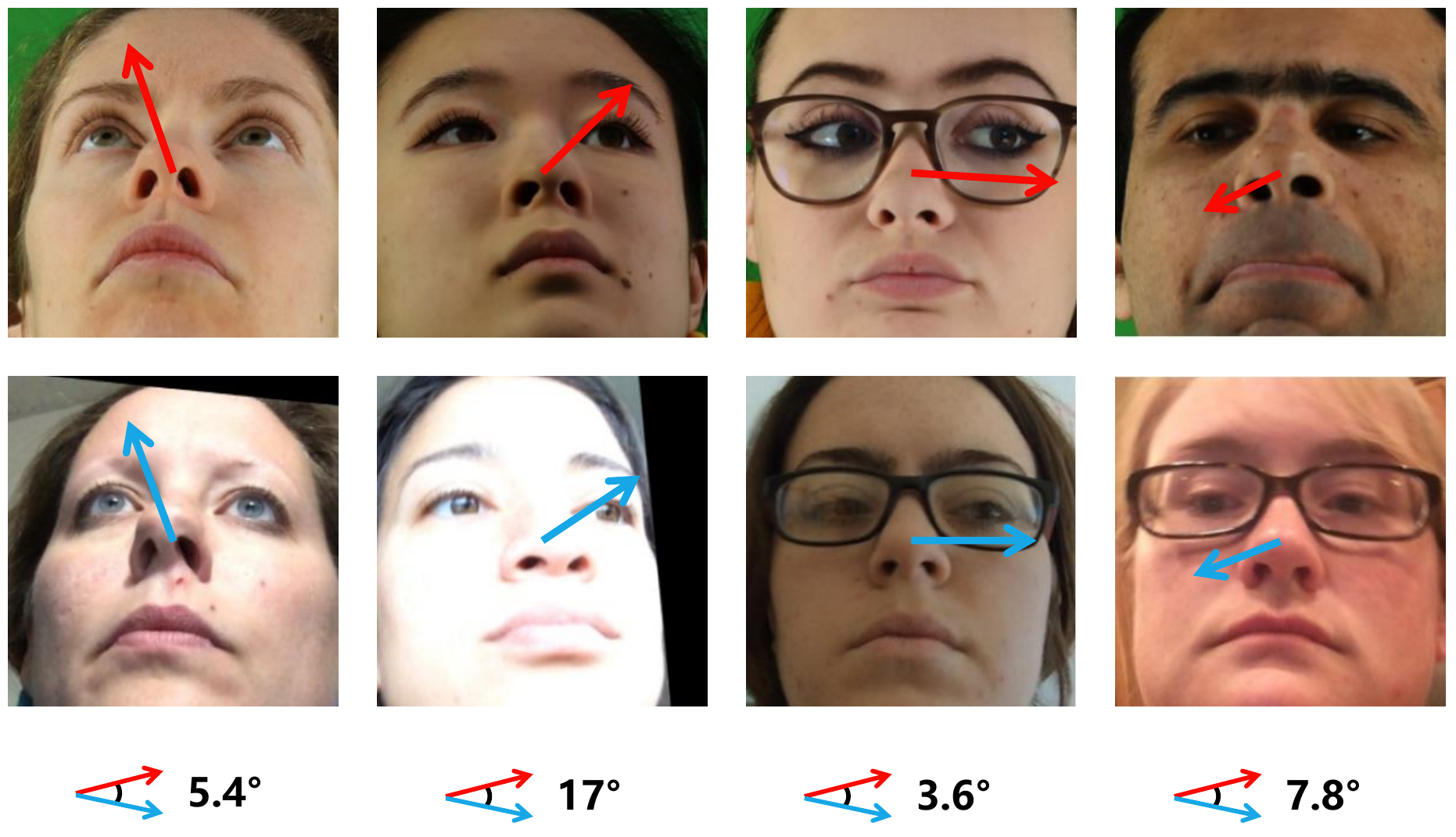}
  \vspace{-0.2cm}
  \caption{The examples of gaze deviation between gaze datasets.
  The red and blue arrows are truth gaze labels. In each column, two images from different datasets have similar head poses and eye rotation angles but significantly different gaze directions.}
  \label{fig:conflict}
  \vspace{-0.3cm}
\end{figure}

Multiple gaze datasets calibrate $(\boldsymbol{R}, \boldsymbol{T})$ through different ways, such as checkerboard, AprilTag \cite{gaze360}, calibration ball \cite{eyediap}, so there are different $\{(\boldsymbol{R}_i^{\prime}, \boldsymbol{T}_i^{\prime})\}^M_{i=1}$, where $\boldsymbol{R}_i^{\prime}$ and $\boldsymbol{T}_i^{\prime}$ are the actual calibration results with calibration error of the $i$-th dataset, and $M$ is the number of datasets.
The calibration error is fixed in a gaze label acquisition system and different among gaze label acquisition systems, which causes the gaze label deviation in different gaze datasets.
Then we analyze the impact of the error in $(\boldsymbol{R}_i^{\prime}, \boldsymbol{T}_i^{\prime})$ on the gaze direction $\boldsymbol{g}$.

\subsection{the Impact of Gaze Label Deviation}
We construct a typical gaze collection environment for analysis (i.e. the distance between gaze origin $\boldsymbol{o}$ and gaze target $t_c$ is 60 cm), the experimental details can be found in the supplementary materials.
In this environment, the $\boldsymbol{o}$ values of different samples remain basically unchanged.
Therefore, we make it a constant $\boldsymbol{C}$ to simplify the problem, so the 3D gaze direction $\boldsymbol{g}^{\prime}=(p^{\prime}_x, p^{\prime}_y, p^{\prime}_z)$ with deviation can be formulated as:
\begin{equation}
\boldsymbol{g}^{\prime} = \frac{ \boldsymbol{R}^{\prime} \boldsymbol{t}_c + \boldsymbol{T}^{\prime} -\boldsymbol{C}}{\left \| \boldsymbol{R}^{\prime} \boldsymbol{t}_c + \boldsymbol{T}^{\prime} -\boldsymbol{C} \right \| }
\nonumber
\end{equation}

\begin{figure*}[htp]
  \centering
  \includegraphics[width=1.0\textwidth]{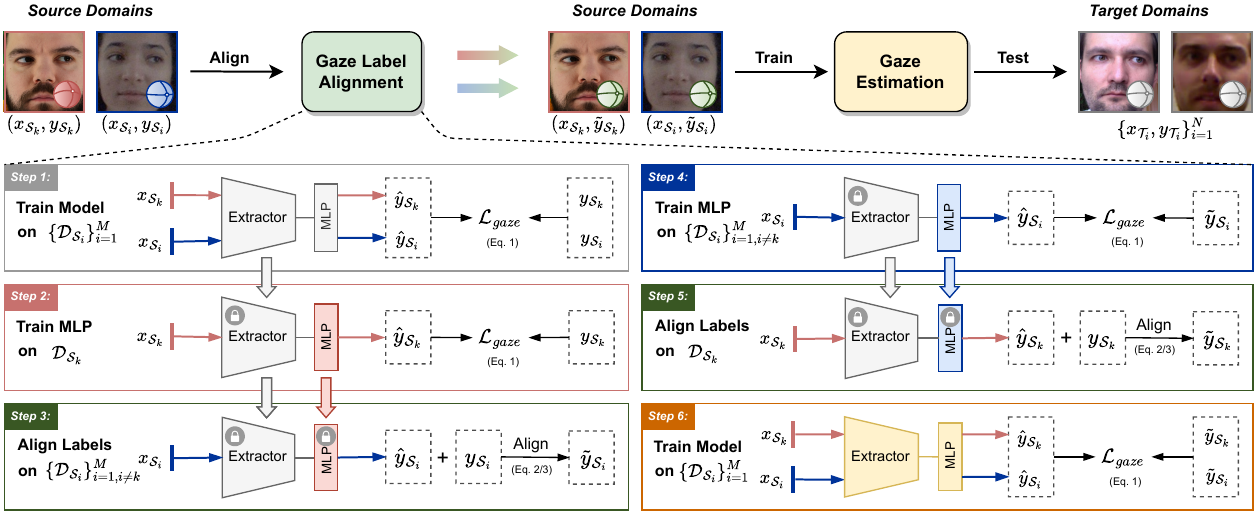}
  \caption{Overview of the proposed GLA method. The GLA is the procedure before  training the gaze estimation model and consists of six steps.
  }
  \label{fig:framework}
  \vspace{-0.6cm}
\end{figure*}

Additionally, the rotation matrix $\boldsymbol{R}^{\prime}$ can be represented by Euler angles $(pitch, yaw, roll)$, the translation matrix $\boldsymbol{T}^{\prime}={\begin{bmatrix} x & y  &z \end{bmatrix}}^T$ denotes the offset on the three coordinate axes.
As a result, there are six variables $(pitch, yaw, roll, x, y, z)$ totally.
We perform sensitivity analysis to analyze the impact of different deviation in $(pitch, yaw, roll, x, y, z)$ on gaze angle separately, then calculate the corresponding gaze deviation.
We assume that the calibration errors of multiple gaze acquisition systems follow a Gaussian distribution $X_1 \sim N\left(\mu, \sigma^2\right)$ on each variable, where $\mu=0$ and $\sigma=1^{\circ}$ or 1cm.
We treat $1^{\circ}$ or 1cm as the unit values for analysis, where $1^{\circ}$ is for $(pitch, yaw, roll)$, and 1cm is for $(x,y,z)$.
So that the 99.7 $\%$ of error is less than 3$^{\circ}$ or 3cm, and we take 3$^{\circ}$ or 3cm as typical calibration error values.
Then, we add a 3$^{\circ}$ or 3cm noise on each variable,
and calculate the gaze deviation $\mathcal{E}_g$ between $\boldsymbol{g}$ and $\boldsymbol{g}^{\prime}$ by:
\begin{equation}
\mathcal{E}_{g}=\arccos \left(\frac{\boldsymbol{g}^{\prime} \cdot \boldsymbol{g}}{\|\boldsymbol{g}^{\prime}\| \|\boldsymbol{g}\|}\right)
  \nonumber
\end{equation}

The results are shown in Fig. \ref{fig:analysis} (a). We notice that the gaze deviation is non-negligible, for example, the deviations caused by $x$ and $y$ are approaching 3${^\circ}$.
In fact, the calibration error is the coupling result of multiple variables, so we add a Gaussian distribution noise $X_2 \sim N\left(v, \tau^2\right)$ on each variable to simulate the real calibration situation, where $v=0$, $\tau=1^{\circ}$ or 1cm.
The Monte Carlo analysis results of gaze deviation within a gaze dataset caused by six variables are displayed in Fig. \ref{fig:analysis} (b), at the unit calibration error values, $\mathcal{E}_g=1.31^{\circ}$.
Further, we change the value of $\tau$. It is found that as the calibration error increases, $\mathcal{E}_g$ also becomes linearly larger.

Actually, we have observed many examples of gaze deviation between multiple gaze datasets, and show them in Fig. \ref{fig:conflict}.
The first row of images come from the same gaze dataset, while the second row of images comes from another dataset.
Two images in the same column have very similar head poses and eye rotation angles, but with a significant deviation in the truth gaze labels.
In a world, the above analysis and phenomenon all illustrate that there are obvious gaze deviations in different gaze datasets.

\subsection{Gaze Label Alignment}
\label{sec:gaze_align}

Fig. \ref{fig:framework} illustrates the overview of our proposed framework. Given multiple source domains, we first use our proposed gaze label alignment (GLA) method to align the gaze labels. Then the multi-domain data with aligned gaze labels are used to train the gaze estimation model. It is worth noting that, our method can be regarded as the pre-process before training gaze estimation models, and it can be combined with any method to further improve its performance.
Suppose we have multiple source domains $\{\mathcal{D}_{S_i}\}_{i=1}^{M}$, the goal of GLA is to align label distribution among these domains.

As shown in Fig. \ref{fig:framework}, the GLA method consists of six steps. We first train on all source domains to eliminate data distribution shift.
After this, we estimate label distribution gap among domains.
If we fix the feature extractor $F$, and train a gaze regressor on a certain domain $\mathcal{D}_{\mathcal{S}_k}$.
Then the predictions of the gaze regressor follow the label distribution of $\mathcal{D}_{\mathcal{S}_k}$.
We use this gaze regressor to predict the gaze directions on other domains $\{\mathcal{D}_{\mathcal{S}_i}\}_{i=1,i\neq k}^{M}$, and get the predicted gaze directions $\hat{y}_{\mathcal{S}_i}$ corresponding to domain $\mathcal{D}_{\mathcal{S}_i}$.
The difference between $\hat{y}_{\mathcal{S}_i}$ and ground truth gaze labels $y_{\mathcal{S}_i}$ can denote the label deviation between $\mathcal{D}_{\mathcal{S}_k}$ and $\mathcal{D}_{\mathcal{S}_i}$.
Thus, for each domain in $\{\mathcal{D}_{\mathcal{S}_i}\}_{i=1,i\neq k}^{M}$ we can fit a function to align $y_{\mathcal{S}_i}$ to $\hat{y}_{\mathcal{S}_i}$, and the aligned label $\tilde{y}_{\mathcal{S}_i}$ can be considered as the new label of $\mathcal{D}_{\mathcal{S}_i}$ without deviation from $\mathcal{D}_{\mathcal{S}_k}$.
Similarly, in other domains $\{\mathcal{D}_{\mathcal{S}_i}\}_{i=1,i\neq k}^{M}$, we train a new regressor with the aligned label $\{\tilde{y}_{\mathcal{S}_i}\}_{i=1,i\neq k}^{M}$ to predict $\hat{y}_{\mathcal{S}_k}$, then calculate the aligned labels $\tilde{y}_{\mathcal{S}_k}$.
Finally, we retrain the entire gaze estimation model on all source domains with aligned labels to furtherly eliminate the label distribution shift among domains.
The detailed description of each step is as follows.

\noindent
\textbf{Step1: Eliminate the data distribution shift.}
Since the distribution shift consists of data distribution $P(X)$ and label distribution $P(Y|X)$, we first need to eliminate data distribution shift. We train gaze estimation model $G=W(F(I))$ on all source domains $\{\mathcal{D}_{\mathcal{S}_i}\}_{i=1}^{M}$, using the convolutional neural networks (CNN) as the feature extractor $F$ and a multi-layer perceptron (MLP) as the gaze regressor $W$.
To avoid the impact of data imbalance, we resample data to maintain the same amount of data for each domain.
The following loss function is used to train the model:
\begin{equation}
\label{eq:l_gaze}
\mathcal{L}_{gaze}\left(\hat{y},y\right)=\arccos \left(\frac{\hat{y} \cdot y}{\|\hat{y}\| \|y\|}\right)
\end{equation}
where $\hat{y}$ is the model predicted gaze direction and $y$ is the ground truth label.

\noindent
\textbf{Step2: Train gaze regressor on selected domain.}
We devide all source domains into two parts, the first part is $\mathcal{D}_{\mathcal{S}_k}$ and the remaining part is $\{\mathcal{D}_{\mathcal{S}_i}\}_{i=1,i\neq k}^{M}$.
Then we fix the the feature extractor $F$ trained on step1 and train a new gaze regressor on $\mathcal{D}_{\mathcal{S}_k}$ using the loss function defined in Eq. \ref{eq:l_gaze}.
In this way, we get a gaze regressor $W_{\mathcal{S}_k}$.

\noindent
\textbf{Step3: Align labels of remaining domains with selected domain.}
First, we use the fixed feature extractor $F$ and the gaze regressor $W_{\mathcal{S}_k}$ as the gaze estimation model, and evaluate it on remaining domains. For each domain $\mathcal{D}_{\mathcal{S}_i}$ in $\{\mathcal{D}_{\mathcal{S}_i}\}_{i=1,i\neq k}^{N}$, we can get the predicted gaze directions $\hat{y}_{S_i}$.
Then, we modify the original truth labels $y_{\mathcal{S}_i}$ to align them with the predicted labels $\hat{y}_{\mathcal{S}_i}$, and generate aligned labels $\tilde{y}_{\mathcal{S}_i}$.
Based on the analysis of gaze label deviation, we believe that the calibration error among datasets mainly comes from the gaze target.
Hence, we use the following alignment function to calculate $\tilde{y}_{\mathcal{S}_i}$:

\begin{equation}
\label{eq:new_label}
\vspace{0.2cm}
y_{\mathcal{S}_i} = \frac{ \boldsymbol{t}_{c_i} -\boldsymbol{o}_{i}}{\left \| \boldsymbol{t}_{c_i} -\boldsymbol{o}_{i} \right \| }
,\;
\tilde{y}_{\mathcal{S}_i} = \frac{ \boldsymbol{\widetilde{R}}_i \boldsymbol{t}_{c_i} + \boldsymbol{\widetilde{T}}_i -\boldsymbol{o}_{i}}{\left \| \boldsymbol{\widetilde{R}}_i \boldsymbol{t}_{c_i} + \boldsymbol{\widetilde{T}}_i -\boldsymbol{o}_{i} \right \| }
\end{equation}

where $\boldsymbol{t}_{c_i}$ is the gaze target in CCS of domain $\mathcal{D}_{\mathcal{S}_i}$, and $\boldsymbol{o}_{i}$ is the corresponding gaze origin.
$\boldsymbol{\widetilde{R}}_i \in \mathbb{R}^{3 \times 3}$ and $\boldsymbol{\widetilde{T}}_i \in \mathbb{R}^{1 \times 3}$ are the fitted parameters of the alignment function, which are shared by all samples and can be estimated by least squares error (LMSE) optimization:

\begin{equation}
\label{eq:align_func}
\min _{\boldsymbol{\widetilde{R}}_i,\boldsymbol{\widetilde{T}}_i} \mathbb{E}_{\left (\tilde{y}_{\mathcal{S}_i}, \hat{y}_{\mathcal{S}_i} \right)} \left [ \mathcal{L}_{gaze}(\tilde{y}_{\mathcal{S}_i}, \hat{y}_{\mathcal{S}_i})\right]
\end{equation}

where $\mathbb{E}$ is the expectation.
Additionally, we can take each person's data as an alignment unit to calculate the corresponding alignment function parameters to eliminate the impact of individual physiological differences.

\noindent
\textbf{Step4: Train gaze regressor on remaining domains.}
We fix the feature extractor $F$ trained on step1 and train a gaze regressor on $\{\mathcal{D}_{\mathcal{S}_i}\}_{i=1,i\neq k}^{N}$ with aligned label $\{\tilde{y}_{\mathcal{S}_i}\}_{i=1,i\neq k}^{N}$.
By this way, we obtain a new gaze regressor $W_{\{\mathcal{D}_{\mathcal{S}_i}\}_{i=1,i\neq k}^{M}}$.

\noindent
\textbf{Step5: Align labels of selected domain with remaining domains.}
In the same way, we use the feature extractor $F$ trained in step1 and the gaze regressor $W_{\{\mathcal{D}_{\mathcal{S}_i}\}_{i=1,i\neq k}^{M}}$ as gaze estimation model,
and evaluate on domain $\mathcal{D}_{\mathcal{S}_k}$.
Then modifying the labels $y_{\mathcal{S}_k}$ according to Eq. \ref{eq:new_label} and Eq. \ref{eq:align_func} to calculate the aligned labels $\tilde{y}_{\mathcal{S}_k}$.

\noindent
\textbf{Step6: Train model with aligned labels.}
Finally, we use the aligned labels $\{\tilde{y}_{\mathcal{S}_i}\}_{i=1}^{M}$ to train gaze feature extractor and regressor from scratch on all source domains, then gain a gaze estimation model that alleviates label domain shifts.

\section{Experiments}
\subsection{Data Preparation}

Due to the wide range of gaze angle, ETH-XGaze \cite{xgaze} and Gaze360 \cite{gaze360} are commonly used as training datasets. Additionally, we add GazeCapture \cite{gazecapture} to construct more source domains.
To maintain consistency with previous works \cite{puregaze, crga, unrega}, we select MPIIFaceGaze \cite{mpii} and EyeDiap \cite{eyediap} as our target datasets.
Respectively, we denote them as $\mathcal{D}_\mathrm{E}$ (ETH-XGaze), $\mathcal{D}_\mathrm{G}$ (Gaze360), $\mathcal{D}_\mathrm{C}$ (GazeCapture), $\mathcal{D}_\mathrm{M}$ (MPIIFaceGaze) and $\mathcal{D}_\mathrm{D}$ (EyeDiap).
For a detailed description, see the supplementary material.

\subsection{Implementation Details}
We employ ResNet-18 \cite{resnet} as our gaze feature extractor $F$, a two layers MLP regressor $W$ to predict gaze angle.
For baseline, we employ SGD Optimizer with a learning rate of $5\times10^{-2}$.
The batch size is 126, and all images are resized to $224\times224$.
We train the gaze estimation model for 30 epochs utilizing a Cosineannealing LR scheduler \cite{sgdr} with a 3 epoch warm-up.
We perform a data augmentation family with a random color field and greyscale like \cite{crga}.

\subsection{Compared Methods}
For baseline, we only use Eq.~\ref{eq:l_gaze} to train the model.
For domain generation (DG) methods, we choose PureGaze \cite{puregaze}, CDG \cite{crga} and CLIP-Gaze \cite{clipgaze} for comparison.
The domain adaptation (DA) methods include PnP-GA \cite{pnpga}, RUDA \cite{ruda}, CSA \cite{crga} and UnReGA \cite{unrega}.
For each method with GLA, we first use baseline method to generate aligned labels $\{\tilde{y}_{\mathcal{S}_i}\}_{i=1}^{M}$, and retrain gaze model with $\{\tilde{y}_{\mathcal{S}_i}\}_{i=1}^{M}$.
More implementation details can be found in the supplementary material.

\subsection{Effectiveness of GLA on Multi-Source Domain}

\begin{table}[tb]
    \renewcommand\arraystretch{1.0}
    \centering
    \setlength{\tabcolsep}{2mm}
    \scalebox{0.90}{
      \begin{tabular}{*{7}{c}}
        \toprule
        \multirow{2}*{\text{Source}} & \multicolumn{2}{c}{$\mathcal{D}_\mathrm{M}$} & \multicolumn{2}{c}{$\mathcal{D}_\mathrm{D}$} & \multicolumn{2}{c}{Avg.} \\
        ~ & {$\emph{w/o}$} & {$\emph{w}$} & {$\emph{w/o}$} & {$\emph{w}$} & {$\emph{w/o}$} & {$\emph{w}$} \\
        \midrule
        {$\mathcal{D}_\mathrm{E}$} & 7.98 & \textbf{6.83} & 7.65 & \textbf{7.38} & 7.82 & \textbf{7.11} \\
        {$\mathcal{D}_\mathrm{C}$} & 6.72 & \textbf{6.61} & 7.91 & \textbf{7.44} & 7.32 & \textbf{7.03} \\
        {$\mathcal{D}_\mathrm{G}$} & 8.02 & \textbf{7.62} & 8.33 & \textbf{7.55} & 8.18 & \textbf{7.59} \\
        \midrule
        $\boldsymbol{\mathcal{D}_\mathrm{E}}$+{$\mathcal{D}_\mathrm{C}$} & 6.83 & \textbf{6.58} & 7.29 & \textbf{6.93} & 7.06 & \textbf{6.76} \\
        $\boldsymbol{\mathcal{D}_\mathrm{E}}$+{$\mathcal{D}_\mathrm{G}$} & 6.77 & \textbf{6.63} & 7.39 & \textbf{7.27} & 7.08 & \textbf{6.95} \\
        $\boldsymbol{\mathcal{D}_\mathrm{G}}$+{$\mathcal{D}_\mathrm{C}$} & 5.42 & \textbf{5.05} & 6.34 & \textbf{5.95} & 5.88 & \textbf{5.50} \\
        $\boldsymbol{\mathcal{D}_\mathrm{E}}$+{$\mathcal{D}_\mathrm{G}$}+{$\mathcal{D}_\mathrm{C}$} & 5.28 & \textbf{4.89} & 5.91 & \textbf{5.71} & 5.60 & \textbf{5.30} \\
      \bottomrule
      \end{tabular}
    }
    \vspace{-0.2cm}
    \caption{Cross-domain evaluation performances of models trained with different source domains.
    Results are reported by angular error ($^{\circ}$), bold denotes the best result on one specific task among each source domain combination.
    Avg. represents the average error of two cross-domain tasks.
    $\emph{w/o}$ indicates without GLA method, and $\emph{w}$ indicates with GLA method.
    The bold dataset means that it is selected as $\mathcal{D}_{\mathcal{S}_k}$.
    }
    \vspace{-0.5cm}
    \label{tab1}
  \end{table}

In order to verify the effectiveness of GLA on multi-source domains training.
We train the baseline models without or with GLA method on different source domains respectively, and conduct cross-domain evaluation.
The training source domains are different combinations of $\mathcal{D}_\mathrm{E}$, $\mathcal{D}_\mathrm{G}$ and $\mathcal{D}_\mathrm{C}$.
When applying GLA method in the single source domain training, we randomly divide the training set into two subsets, and then perform  GLA operation.
The test datasets are $\mathcal{D}_\mathrm{M}$ and $\mathcal{D}_\mathrm{D}$, and the evaluation results are reported in Table \ref{tab1}.
Note that the bold dataset means that it is selected as $\mathcal{D}_{\mathcal{S}_k}$.

We can see that, the method with GLA outperforms the method without GLA regardless of on single training domain, two training domains or three training domains. On single training domain, GLA can eliminate the individual deviation, and on multiple domains, GLA can not only eliminate individual deviation but also the deviation of data acquisition systems among different domains.

It is worth noting that the model without GLA trained on $\boldsymbol{\mathcal{D}_\mathrm{E}}$ $+\mathcal{D}_\mathrm{C}$ has a higher error on $\mathcal{D}_\mathrm{M}$ than the model trained on $\mathcal{D}_\mathrm{C}$ (6.83$^{\circ}$ vs 6.72$^{\circ}$).
We conjecture that the label gap between $\mathcal{D}_\mathrm{E}$ and $\mathcal{D}_\mathrm{C}$ has led to a higher test error. When we apply GLA in training to eliminate the label shift, the error of the model trained with $\boldsymbol{\mathcal{D}_\mathrm{E}}$ $+\mathcal{D}_\mathrm{C}$ becomes lower than the model trained with $\mathcal{D}_\mathrm{C}$ (6.58$^{\circ}$ vs 6.61$^{\circ}$). With GLA, it is guaranteed that the multi-source domains trained models are always better than the single domain trained models, and more training domains can further improve the performance.

\subsection{Plugging into the SOTA Methods}

\begin{table}[tb]
    \renewcommand\arraystretch{1.0}
    \centering
    \setlength{\tabcolsep}{1.5mm}
    \scalebox{0.90}{
        \begin{tabular}{*{8}{c}}
            \toprule
            \multirow{2}*{\text{Task}} & \multirow{2}*{\text{Methods}} & \multicolumn{2}{c}{$\mathcal{D}_\mathrm{M}$} & \multicolumn{2}{c}{$\mathcal{D}_\mathrm{D}$} & \multicolumn{2}{c}{Avg.} \\
            ~ & ~ & {$\emph{w/o}$} & {$\emph{w}$} & {$\emph{w/o}$} & {$\emph{w}$} & {$\emph{w/o}$} & {$\emph{w}$} \\
            \midrule
            \multirow{4}*{\text{DG}} & Baseline & 6.83 & \textbf{6.58} & 7.39 & \textbf{6.93} & 7.11 & \textbf{6.75} \\
            ~ & PureGaze & 6.42 & \textbf{6.28} & 6.77 & \textbf{6.34} & 6.59 & \textbf{6.31} \\
            ~ & CDG & 6.63 & \textbf{6.34} & 6.67 & \textbf{6.42} & 6.65 & \textbf{6.38} \\
            ~ & CLIP-Gaze & 6.36 & \textbf{6.13} & 6.45 & \textbf{6.30} & 6.40 & \textbf{6.22} \\
            \midrule
            \multirow{4}*{\text{UDA}} & PnP-GA & 6.24 & \textbf{5.77} & 6.32 & \textbf{6.22} & 6.28 & \textbf{6.00} \\
            ~ & RUDA & 6.03 & \textbf{5.60} & 6.07 & \textbf{5.75} & 6.05 & \textbf{5.68} \\
            ~ & CSA & 5.79 & \textbf{5.58} & 6.14 & \textbf{6.00} & 5.97 & \textbf{5.79} \\
            ~ & UnReGA & 5.27 & \textbf{5.05} & 5.86 & \textbf{5.71} & 5.56 & \textbf{5.38} \\
            \midrule
            {\text{SDA}} & Finetuning & 5.17 & \textbf{4.62} & 5.69 & \textbf{5.40} & 5.43 & \textbf{5.01} \\
          \bottomrule
          \end{tabular}
    }
    \vspace{-0.2cm}
    \caption{Performances of GLA method embedded into SOTA methods.
    }
    \vspace{-0.5cm}
    \label{tab2}
\end{table}

Our proposed GLA method can reduce the label gap between different source domains and generate the aligned gaze labels, which is complementary with current methods.
By training on the aligned labels instead of the original gaze labels, we can plug the GLA method into SOTA methods to further improve the performance of gaze estimation model.
Specifically, we use $\boldsymbol{\mathcal{D}_\mathrm{E}}$ $+\mathcal{D}_\mathrm{C}$ as source domains, and generate aligned gaze labels through the baseline method with GLA.
Then, we train various SOTA methods on the aligned labels,
and evaluate the models on $\mathcal{D}_\mathrm{M}$ and $\mathcal{D}_\mathrm{D}$. The experimental results are shown in Table \ref{tab2}.

For domain generalization (DG) tasks, we compare the performance of PureGaze, CDG, and CLIP-Gaze on the original labels ($\emph{w/o}$ GLA) and aligned labels ($\emph{w}$ GLA).
It can be seen that each DG method ($\emph{w/o}$ GLA) performs better than the baseline method as they reduce the deviation between data distributions. GLA can further decrease the errors on the basis of DG methods, as it can further eliminate the deviation between label distributions.
For domain adaptation tasks, we train unsupervised domain adaptation (UDA) methods PnP-GA, RUDA, CSA and UnReGA, and supervised domain adaptation (SDA) method with the original labels ($\emph{w/o}$ GLA) or aligned labels ($\emph{w}$ GLA) on the source domains.
We randomly choose 100 samples from target domain and report average results of 20 repeated trials.
It can be seen that plugging GLA can further reduce estimation errors, achieving a relative improvement of 3-8$\%$, specifically 5$\%$ (on PnP-GA), 6$\%$ (on RUDA), 3$\%$ (on CSA), 3$\%$ (on UnReGA), and 8$\%$ (on fine-tuning).

The above experiments demonstrate that GLA can be utilized with DG and DA methods to further improve the performance of gaze estimation, which indicates the GLA method is a general and effective method.

\subsection{Ablation Study}
\subsubsection{Ablation Study on Alignment Functions}
\begin{table}[tb]
  \renewcommand\arraystretch{1.0}
  \centering
  \setlength{\tabcolsep}{3mm}
  \scalebox{0.9}{
    \begin{tabular}{*{4}{c}}
      \toprule
      {\text{Methods}} & {$\mathcal{D}_\mathrm{M}$} & {$\mathcal{D}_\mathrm{D}$} & {Avg.} \\
      \midrule
      Baseline & 6.83 & 7.39 & 7.11 \\
      \midrule
      {GLA$_{offset}$} & 6.78 & 7.11 & 6.95 \\
      {GLA$_{linear}$} & 6.69 & 7.04 & 6.87 \\
      {GLA$_{affine}$} & 6.65 & \underline{6.97} & 6.81 \\
      {GLA$_{homography}$} & \underline{6.60} & 6.99 & \underline{6.80} \\
      {GLA$_{RT}$} & \textbf{6.58} & \textbf{6.93} & \textbf{6.76} \\
    \bottomrule
    \end{tabular}
  }
    \vspace{-0.2cm}
  \caption{Performances comparison of different multi-source domain alignment methods.}
    \vspace{-0.5cm}
  \label{tab3}
\end{table}

When performing label alignment, we can use different alignment functions to calculate the aligned labels, so we conduct ablation study on alignment functions, these GLA method variants are described as follows.
\begin{itemize}
\item $\text{GLA}_{offset}$: calculating the offset between $\hat{y}$ and $y$, and mapping the original labels $y$ to aligned labels $\tilde{y}$.
The alignment function is: $\tilde{y} = y + \boldsymbol{B}$, where $y=(\theta, \phi)$ is the polar equivalent representation of gaze direction $\boldsymbol{g}$. $\boldsymbol{B}\in \mathbb{R}^{1 \times 2}$ is a constant.
\item $\text{GLA}_{linear}$: it corresponds the alignment function is: $\tilde{y} = \boldsymbol{K} \cdot y + \boldsymbol{B}$, where $\boldsymbol{K} \in \mathbb{R}^{1 \times 2}$ and $\boldsymbol{B} \in \mathbb{R}^{1 \times 2}$ are linear parameters.
\item $\text{GLA}_{affine}$: it corresponds the alignment function is: $\tilde{y} = \boldsymbol{A} \cdot y + \boldsymbol{T}$, where $\boldsymbol{A} \in \mathbb{R}^{2 \times 2}$ and $\boldsymbol{T} \in \mathbb{R}^{1 \times 2}$ are affine transformation parameters.
\item $\text{GLA}_{homography}$: it corresponds the alignment function is: $[\tilde{\theta},\tilde{\phi},1]^T \;=\; \boldsymbol{H}\cdot [\theta,\phi,1]^T$.
where $\boldsymbol{H} \in \mathbb{R}^{3 \times 3}$ is homography transformation parameters.
\item $\text{GLA}_{RT}$: it represents the method using Eq. \ref{eq:new_label}.
\end{itemize}

We use $\boldsymbol{\mathcal{D}_\mathrm{E}}$ $+\mathcal{D}_\mathrm{C}$ as source domains.
The experimental results are shown in Table \ref{tab3}.
It can be seen that various alignment functions can achieve good performance improvement compared with the baseline without GLA.
among which $\text{GLA}_{RT}$ is better than other alignment methods.
This also proves that our analysis of gaze label deviation and definition of alignment function are correct and effective.

\subsubsection{Ablation Study on Alignment Units}

\begin{table}[tb]
  \renewcommand\arraystretch{1.0}
      \centering
      \setlength{\tabcolsep}{0.8mm}
      \scalebox{0.87}{
          \begin{tabular}{*{10}{c}}
          \toprule
          \multirow{2}*{\text{Source}} & \multicolumn{3}{c}{$\mathcal{D}_\mathrm{M}$} & \multicolumn{3}{c}{$\mathcal{D}_\mathrm{D}$} & \multicolumn{3}{c}{Avg.} \\
          ~ & {$\emph{base.}$} & {$\emph{pers.}$} & {$\emph{ds.}$} & {$\emph{base.}$} & {$\emph{pers.}$} & {$\emph{ds.}$} & {$\emph{base.}$} & {$\emph{pers.}$} & {$\emph{ds.}$} \\
          \midrule
          $\boldsymbol{\mathcal{D}_\mathrm{E}}$+{$\mathcal{D}_\mathrm{C}$} & 6.83 & \textbf{6.58} & \underline{6.75} & 7.29 & \textbf{6.93} & \underline{7.17} & 7.06 & \textbf{6.76} & \underline{6.96} \\
          $\boldsymbol{\mathcal{D}_\mathrm{E}}$+{$\mathcal{D}_\mathrm{G}$} & 6.77 & \textbf{6.63} & \underline{6.70} & 7.39 & \textbf{7.27} & \underline{7.30} & 7.08 & \textbf{6.95} & \underline{7.00} \\
          $\boldsymbol{\mathcal{D}_\mathrm{G}}$+{$\mathcal{D}_\mathrm{C}$} & 5.42 & \textbf{5.05} & \underline{5.31} & 6.34 & \textbf{5.95} & \underline{6.13} & 5.88 & \textbf{5.50} & \underline{5.72} \\
          $\boldsymbol{\mathcal{D}_\mathrm{E}}$+{$\mathcal{D}_\mathrm{G}$}+{$\mathcal{D}_\mathrm{C}$} & 5.28 & \textbf{4.89} & \underline{5.14} & 5.91 & \textbf{5.71} & \underline{5.82} & 5.60 & \textbf{5.30} & \underline{5.48} \\
          \bottomrule
          \end{tabular}
      }
    \vspace{-0.1cm}
  \caption{Performance comparison of different label alignment units. Bold and underline denotes the best and the second best result in a specific task among each source domain combination.}
    \vspace{-0.3cm}
  \label{tab4}
\end{table}

\begin{table}[tb]
  \renewcommand\arraystretch{1.0}
      \centering
      \setlength{\tabcolsep}{1.5mm}
      \scalebox{0.95}{
          \begin{tabular}{*{6}{c}}
          \toprule
          {\text{Methods}} & {$\mathcal{D}_{\mathcal{S}_k}$} & {$\{\mathcal{D}_{\mathcal{S}_i}\}_{i=1,i\neq k}^{M}$} & {$\mathcal{D}_\mathrm{M}$} & {$\mathcal{D}_\mathrm{D}$} & {Avg.} \\
          \midrule
          {Baseline} & - & - & 6.83 & 7.39 & 7.11 \\
          \midrule
          \multirow{3}*{{$\mathcal{D}_\mathrm{E}$}+{$\mathcal{D}_\mathrm{G}$}+{$\mathcal{D}_\mathrm{C}$}} & {$\mathcal{D}_\mathrm{E}$} & {$\mathcal{D}_\mathrm{G}$}+{$\mathcal{D}_\mathrm{C}$} & \underline{4.89} & \textbf{5.71} & \textbf{5.30} \\
          ~ & {$\mathcal{D}_\mathrm{G}$} & {$\mathcal{D}_\mathrm{E}$}+{$\mathcal{D}_\mathrm{C}$} & \textbf{4.84} & 5.82 & \underline{5.33} \\
          ~ & {$\mathcal{D}_\mathrm{C}$} & {$\mathcal{D}_\mathrm{E}$}+{$\mathcal{D}_\mathrm{G}$} & 5.17 & \underline{5.79} & 5.48 \\
          \bottomrule
          \end{tabular}
      }
    \vspace{-0.1cm}
  \caption{Performances comparison of different source domain as $\mathcal{D}_{\mathcal{S}_k}$.}
    \vspace{-0.7cm}
  \label{tab5}
\end{table}

We conduct comparative experiments to explore appropriate alignment basic units.
Specifically, we treat the data from the same person as a alignment unit or the whole dataset as a unit to align.
$\boldsymbol{\mathcal{D}_\mathrm{E}}$ $+\mathcal{D}_\mathrm{C}$ are training domains.
The experimental results are shown in Table \ref{tab4}.
$\emph{base.}$ in the table represents the baseline method without GLA, $\emph{pers.}$ denotes treating each person's data as a single domain to align, and $\emph{ds.}$ denotes taking each dataset as an alignment unit.
The results show that for each combination of source domains, the method with alignment on dataset level achieves better performance than the one without alignment, and alignment on the person level performs the best. It demonstrates that there exist both systematic deviation between different data acquisition systems and individual deviation between different persons simultaneously.
Therefore, it is better to treat the personal data as a separate alignment unit when applying GLA.

\subsubsection{Ablation Study on $\mathcal{D}_{\mathcal{S}_k}$ Selection}

When there are multiple source domains, we can select different domain as $\mathcal{D}_{\mathcal{S}_k}$.
Therefore, we use $\mathcal{D}_\mathrm{E}+\mathcal{D}_\mathrm{G}+\mathcal{D}_\mathrm{C}$ as source domains, and treat $\mathcal{D}_\mathrm{E}$, $\mathcal{D}_\mathrm{G}$ and $\mathcal{D}_\mathrm{C}$ as $\mathcal{D}_{\mathcal{S}_k}$ to align labels respectively.

The experimental results are shown in Table \ref{tab5}.
The test errors of GLA training models with different domain as $\mathcal{D}_{\mathcal{S}_k}$ are more than 1.63$^{\circ}$ lower than the baseline training models.
The error difference between different $\mathcal{D}_{\mathcal{S}_k}$ is within 0.18$^{\circ}$, which is much smaller than the performance benefits brought by GLA (0.18$^{\circ}$ vs 1.63$^{\circ}$).
Consequently, although different $\mathcal{D}_{\mathcal{S}_k}$ may result in slight performance changes, GLA can still bring stable performance improvement.

\subsubsection{Ablation Study on Alignment Times}

We retrain the model with aligned labels to reduce the domain gap of label distribution between multiple source domains.
Therefore, we explore the impact of different label alignment times on model performance.
The process described in Fig. \ref{fig:framework} is referred to as one complete alignment.
When GLA is performed multiple times, each complete alignment results in aligned labels, which are used as the original labels for the next alignment operation.
We use $\boldsymbol{\mathcal{D}_\mathrm{E}}$ $+\mathcal{D}_\mathrm{C}$ as source domains, and test on $\mathcal{D}_\mathrm{M}$ and $\mathcal{D}_\mathrm{D}$.
The experimental results are shown in Table \ref{tab6}. It can be seen that increasing the number of alignment times does not bring a significant improvement in performance (the average errors are 6.76$^{\circ}$, 6.72$^{\circ}$ and 6.74$^{\circ}$ respectively).
Therefore, one alignment is enough.

\begin{table}[tb]
    \renewcommand\arraystretch{1.0}
    \centering
    \setlength{\tabcolsep}{3mm}
    \scalebox{0.9}{
      \begin{tabular}{*{4}{c}}
        \toprule
        {\text{Times}} & {$\mathcal{D}_\mathrm{M}$} & {$\mathcal{D}_\mathrm{D}$} & {Avg.} \\
        \midrule
        Baseline & 6.83 & 7.39 & 7.11 \\
        \midrule
        {GLA $\times$ 1} & \textbf{6.58} & 6.93 & 6.76 \\
        {GLA $\times$ 2} & 6.60 & \textbf{6.89} & \textbf{6.72} \\
        {GLA $\times$ 3} & \underline{6.59} & \underline{6.90} & \underline{6.74} \\
      \bottomrule
      \end{tabular}
    }
    \vspace{-0.1cm}
    \caption{Performances comparison of different label alignment times.}
    \vspace{-0.5cm}
    \label{tab6}
\end{table}

\begin{figure}[tb]
        \centering
        \includegraphics[width=0.425\textwidth]{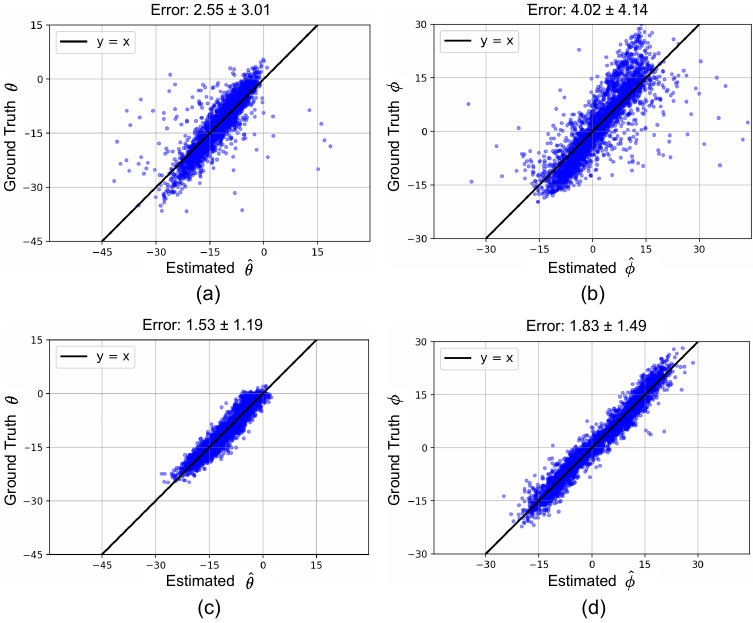}
        \vspace{-0.2cm}
        \caption{The scatter plot of estimated $\hat{\phi}$ (X-axis) and ground truth $\phi$ (Y-axis) of different models on a specific person of $\mathcal{D}_\mathrm{M}$. (a) and (b) are the results of the model without GLA. (c) and (d) are the results of the model with GLA.}
        \label{fig:visualize}
        \vspace{-0.6cm}
    \end{figure}

\subsection{Visualization Analysis}
\label{sec:visual}

We visualize the scatter plot of the estimated $\hat{\theta}, \hat{\phi}$ (X-axis) and ground truth $\theta,\phi$ (Y-axis) of different models in Fig. \ref{fig:visualize}.
The first row corresponds to the results of models $\emph{w/o}$ GLA, and the second row shows the results of models $\emph{w}$ GLA respectively.
All models are trained on $\boldsymbol{\mathcal{D}_\mathrm{E}}$+{$\mathcal{D}_\mathrm{G}$}+{$\mathcal{D}_\mathrm{C}$} and predict on a specific person of target domain $\mathcal{D}_\mathrm{M}$.
We can find that the scatters of models $\emph{w/o}$ GLA are more divergent, and the scatters of models $\emph{w}$ GLA are more clustered.
At the same time, models $\emph{w/o}$ GLA have higher errors than models $\emph{w}$ GLA ($\theta$: 2.55$^{\circ}$ vs 1.53$^{\circ}$, $\phi$: 4.02$^{\circ}$ vs 1.83$^{\circ}$).
This shows GLA can effectively alleviate the label distribution shift and reduce the model prediction uncertainty.

\section{Conclusion}
In this paper, we have identified the label distribution shift as a significant factor that affects the generalization error in gaze estimation. To address this issue, we have proposed the Gaze Label Alignment (GLA) algorithm, which aims to align the label distributions among different domains. It is easy to combine GLA with SOTA gaze estimation methods.
Through extensive experiments, we have demonstrated the effectiveness of our GLA method. When combined with SOTA domain generalization and adaptation methods, it can also lead to significant improvements in performance.

\section{Acknowledgements}
This work was sponsored by National Key R$\&$D Program of China (2023YFE0204200).

\bibliography{aaai25}

\begin{thebibliography}{37}
\providecommand{\natexlab}[1]{#1}

\bibitem[{Admoni and Scassellati(2017)}]{admoni2017social}
Admoni, H.; and Scassellati, B. 2017.
\newblock Social eye gaze in human-robot interaction: a review.
\newblock \emph{Journal of Human-Robot Interaction}, 6(1): 25--63.

\bibitem[{Bao et~al.(2022)Bao, Liu, Wang, and Lu}]{ruda}
Bao, Y.; Liu, Y.; Wang, H.; and Lu, F. 2022.
\newblock Generalizing Gaze Estimation With Rotation Consistency.
\newblock In \emph{Proceedings of the IEEE/CVF Conference on Computer Vision
  and Pattern Recognition}, 4207--4216.

\bibitem[{Biswas et~al.(2021)}]{attention}
Biswas, P.; et~al. 2021.
\newblock Appearance-based gaze estimation using attention and difference
  mechanism.
\newblock In \emph{Proceedings of the IEEE/CVF conference on computer vision
  and pattern recognition}, 3143--3152.

\bibitem[{Cai et~al.(2023)Cai, Zeng, Shan, and Chen}]{unrega}
Cai, X.; Zeng, J.; Shan, S.; and Chen, X. 2023.
\newblock Source-Free Adaptive Gaze Estimation by Uncertainty Reduction.
\newblock In \emph{Proceedings of the IEEE/CVF Conference on Computer Vision
  and Pattern Recognition}, 22035--22045.

\bibitem[{Chang et~al.(2019)Chang, Matias Di~Martino, Qiu, Espinosa, and
  Sapiro}]{chang2019salgaze}
Chang, Z.; Matias Di~Martino, J.; Qiu, Q.; Espinosa, S.; and Sapiro, G. 2019.
\newblock Salgaze: Personalizing gaze estimation using visual saliency.
\newblock In \emph{Proceedings of the IEEE/CVF International Conference on
  Computer Vision Workshops}, 0--0.

\bibitem[{Cheng and Bao(2022)}]{puregaze}
Cheng, Y.; and Bao, Y. 2022.
\newblock Puregaze: Purifying gaze feature for generalizable gaze estimation.
\newblock In \emph{Proceedings of the AAAI Conference on Artificial
  Intelligence}, volume~36, 436--443.

\bibitem[{Cheng et~al.(2020)Cheng, Huang, Wang, Qian, and Lu}]{coarsetofine}
Cheng, Y.; Huang, S.; Wang, F.; Qian, C.; and Lu, F. 2020.
\newblock A coarse-to-fine adaptive network for appearance-based gaze
  estimation.
\newblock In \emph{Proceedings of the AAAI conference on artificial
  intelligence}, volume~34, 10623--10630.

\bibitem[{Cheng, Lu, and Zhang(2018)}]{Cheng_2018_ECCV}
Cheng, Y.; Lu, F.; and Zhang, X. 2018.
\newblock Appearance-Based Gaze Estimation via Evaluation-Guided Asymmetric
  Regression.
\newblock In \emph{Proceedings of the European Conference on Computer Vision
  (ECCV)}.

\bibitem[{Funes~Mora, Monay, and Odobez(2014)}]{eyediap}
Funes~Mora, K.~A.; Monay, F.; and Odobez, J.-M. 2014.
\newblock Eyediap: A database for the development and evaluation of gaze
  estimation algorithms from rgb and rgb-d cameras.
\newblock In \emph{Proceedings of the symposium on eye tracking research and
  applications}, 255--258.

\bibitem[{Guestrin and Eizenman(2006)}]{pupilcenter}
Guestrin, E.~D.; and Eizenman, M. 2006.
\newblock General theory of remote gaze estimation using the pupil center and
  corneal reflections.
\newblock \emph{IEEE Transactions on biomedical engineering}, 53(6):
  1124--1133.

\bibitem[{Guo et~al.(2020)Guo, Yuan, Zhang, Chi, Ling, and Zhang}]{dagen}
Guo, Z.; Yuan, Z.; Zhang, C.; Chi, W.; Ling, Y.; and Zhang, S. 2020.
\newblock Domain adaptation gaze estimation by embedding with prediction
  consistency.
\newblock In \emph{Proceedings of the Asian Conference on Computer Vision}.

\bibitem[{He et~al.(2016)He, Zhang, Ren, and Sun}]{resnet}
He, K.; Zhang, X.; Ren, S.; and Sun, J. 2016.
\newblock Deep residual learning for image recognition.
\newblock In \emph{Proceedings of the IEEE conference on computer vision and
  pattern recognition}, 770--778.

\bibitem[{Kellnhofer et~al.(2019)Kellnhofer, Recasens, Stent, Matusik, and
  Torralba}]{gaze360}
Kellnhofer, P.; Recasens, A.; Stent, S.; Matusik, W.; and Torralba, A. 2019.
\newblock Gaze360: Physically unconstrained gaze estimation in the wild.
\newblock In \emph{Proceedings of the IEEE/CVF International Conference on
  Computer Vision}, 6912--6921.

\bibitem[{Krafka et~al.(2016)Krafka, Khosla, Kellnhofer, Kannan, Bhandarkar,
  Matusik, and Torralba}]{gazecapture}
Krafka, K.; Khosla, A.; Kellnhofer, P.; Kannan, H.; Bhandarkar, S.; Matusik,
  W.; and Torralba, A. 2016.
\newblock Eye tracking for everyone.
\newblock In \emph{Proceedings of the IEEE conference on computer vision and
  pattern recognition}, 2176--2184.

\bibitem[{Li et~al.(2020)Li, Wang, Wan, Wang, Li, and Kot}]{li2020domain}
Li, H.; Wang, Y.; Wan, R.; Wang, S.; Li, T.-Q.; and Kot, A. 2020.
\newblock Domain generalization for medical imaging classification with
  linear-dependency regularization.
\newblock \emph{Advances in neural information processing systems}, 33:
  3118--3129.

\bibitem[{Li et~al.(2018)Li, Tian, Gong, Liu, Liu, Zhang, and Tao}]{li2018deep}
Li, Y.; Tian, X.; Gong, M.; Liu, Y.; Liu, T.; Zhang, K.; and Tao, D. 2018.
\newblock Deep domain generalization via conditional invariant adversarial
  networks.
\newblock In \emph{Proceedings of the European conference on computer vision
  (ECCV)}, 624--639.

\bibitem[{Liu et~al.(2019)Liu, Yu, Mora, and Odobez}]{liu2019differential}
Liu, G.; Yu, Y.; Mora, K. A.~F.; and Odobez, J.-M. 2019.
\newblock A differential approach for gaze estimation.
\newblock \emph{IEEE transactions on pattern analysis and machine
  intelligence}, 43(3): 1092--1099.

\bibitem[{Liu et~al.(2021{\natexlab{a}})Liu, Nie, Zhang, and
  Li}]{liu2021anisotropic}
Liu, H.; Nie, H.; Zhang, Z.; and Li, Y.-F. 2021{\natexlab{a}}.
\newblock Anisotropic angle distribution learning for head pose estimation and
  attention understanding in human-computer interaction.
\newblock \emph{Neurocomputing}, 433: 310--322.

\bibitem[{Liu et~al.(2021{\natexlab{b}})Liu, Liu, Wang, and Lu}]{pnpga}
Liu, Y.; Liu, R.; Wang, H.; and Lu, F. 2021{\natexlab{b}}.
\newblock Generalizing Gaze Estimation with Outlier-guided Collaborative
  Adaptation.
\newblock In \emph{Proceedings of the IEEE/CVF International Conference on
  Computer Vision}.

\bibitem[{Loshchilov and Hutter(2016)}]{sgdr}
Loshchilov, I.; and Hutter, F. 2016.
\newblock Sgdr: Stochastic gradient descent with warm restarts.
\newblock \emph{arXiv preprint arXiv:1608.03983}.

\bibitem[{Motiian et~al.(2017)Motiian, Piccirilli, Adjeroh, and
  Doretto}]{motiian2017unified}
Motiian, S.; Piccirilli, M.; Adjeroh, D.~A.; and Doretto, G. 2017.
\newblock Unified deep supervised domain adaptation and generalization.
\newblock In \emph{Proceedings of the IEEE international conference on computer
  vision}, 5715--5725.

\bibitem[{Murphy-Chutorian and Trivedi(2010)}]{murphy2010head}
Murphy-Chutorian, E.; and Trivedi, M.~M. 2010.
\newblock Head pose estimation and augmented reality tracking: An integrated
  system and evaluation for monitoring driver awareness.
\newblock \emph{IEEE Transactions on intelligent transportation systems},
  11(2): 300--311.

\bibitem[{Shah et~al.(2022)Shah, Sun, Zaman, Hussain, Shoaib, and
  Pei}]{shah2022driver}
Shah, S.~M.; Sun, Z.; Zaman, K.; Hussain, A.; Shoaib, M.; and Pei, L. 2022.
\newblock A driver gaze estimation method based on deep learning.
\newblock \emph{Sensors}, 22(10): 3959.

\bibitem[{Shrivastava et~al.(2017)Shrivastava, Pfister, Tuzel, Susskind, Wang,
  and Webb}]{shrivastava2017learning}
Shrivastava, A.; Pfister, T.; Tuzel, O.; Susskind, J.; Wang, W.; and Webb, R.
  2017.
\newblock Learning from simulated and unsupervised images through adversarial
  training.
\newblock In \emph{Proceedings of the IEEE conference on computer vision and
  pattern recognition}, 2107--2116.

\bibitem[{Sugano, Matsushita, and Sato(2012)}]{sugano2012appearance}
Sugano, Y.; Matsushita, Y.; and Sato, Y. 2012.
\newblock Appearance-based gaze estimation using visual saliency.
\newblock \emph{IEEE transactions on pattern analysis and machine
  intelligence}, 35(2): 329--341.

\bibitem[{Vicente et~al.(2015)Vicente, Huang, Xiong, De~la Torre, Zhang, and
  Levi}]{vicente2015driver}
Vicente, F.; Huang, Z.; Xiong, X.; De~la Torre, F.; Zhang, W.; and Levi, D.
  2015.
\newblock Driver gaze tracking and eyes off the road detection system.
\newblock \emph{IEEE Transactions on Intelligent Transportation Systems},
  16(4): 2014--2027.

\bibitem[{Wang et~al.(2022{\natexlab{a}})Wang, Lan, Liu, Ouyang, Qin, Lu, Chen,
  Zeng, and Yu}]{wang2022generalizing}
Wang, J.; Lan, C.; Liu, C.; Ouyang, Y.; Qin, T.; Lu, W.; Chen, Y.; Zeng, W.;
  and Yu, P. 2022{\natexlab{a}}.
\newblock Generalizing to unseen domains: A survey on domain generalization.
\newblock \emph{IEEE Transactions on Knowledge and Data Engineering}.

\bibitem[{Wang et~al.(2022{\natexlab{b}})Wang, Jiang, Li, Ni, Dai, Li, Xiong,
  and Li}]{crga}
Wang, Y.; Jiang, Y.; Li, J.; Ni, B.; Dai, W.; Li, C.; Xiong, H.; and Li, T.
  2022{\natexlab{b}}.
\newblock Contrastive Regression for Domain Adaptation on Gaze Estimation.
\newblock In \emph{2022 IEEE/CVF Conference on Computer Vision and Pattern
  Recognition (CVPR)}, 19354--19363.

\bibitem[{Xu, Wang, and Lu(2023)}]{gazeconsistence}
Xu, M.; Wang, H.; and Lu, F. 2023.
\newblock Learning a Generalized Gaze Estimator from Gaze-Consistent Feature.
\newblock \emph{Proceedings of the AAAI Conference on Artificial Intelligence},
  37(3): 3027--3035.

\bibitem[{Yin et~al.(2024{\natexlab{a}})Yin, Wang, Dai, and Wu}]{yin2024nerf}
Yin, P.; Wang, J.; Dai, J.; and Wu, X. 2024{\natexlab{a}}.
\newblock Nerf-gaze: A head-eye redirection parametric model for gaze
  estimation.
\newblock In \emph{Proceedings of the IEEE International Conference on
  Acoustics, Speech and Signal Processing (ICASSP)}, 2760--2764. IEEE.

\bibitem[{Yin et~al.(2024{\natexlab{b}})Yin, Wang, Zeng, Xie, and
  Zhu}]{yin2025lg}
Yin, P.; Wang, J.; Zeng, G.; Xie, D.; and Zhu, J. 2024{\natexlab{b}}.
\newblock LG-Gaze: Learning Geometry-aware Continuous Prompts for
  Language-Guided Gaze Estimation.
\newblock In \emph{Proceedings of the European conference on computer vision
  (ECCV)}, 1--17. Springer.

\bibitem[{Yin et~al.(2024{\natexlab{c}})Yin, Zeng, Wang, and Xie}]{clipgaze}
Yin, P.; Zeng, G.; Wang, J.; and Xie, D. 2024{\natexlab{c}}.
\newblock CLIP-Gaze: Towards General Gaze Estimation via Visual-Linguistic
  Model.
\newblock In \emph{Proceedings of the AAAI Conference on Artificial
  Intelligence}, volume~38, 6729--6737.

\bibitem[{Yu, Liu, and Odobez(2018)}]{multitask}
Yu, Y.; Liu, G.; and Odobez, J.-M. 2018.
\newblock Deep multitask gaze estimation with a constrained landmark-gaze
  model.
\newblock In \emph{Proceedings of the European conference on computer vision
  (ECCV) workshops}, 0--0.

\bibitem[{Zhang, Yao, and Cai(2018)}]{zhang2018efficient}
Zhang, C.; Yao, R.; and Cai, J. 2018.
\newblock Efficient eye typing with 9-direction gaze estimation.
\newblock \emph{Multimedia Tools and Applications}, 77: 19679--19696.

\bibitem[{Zhang et~al.(2020)Zhang, Park, Beeler, Bradley, Tang, and
  Hilliges}]{xgaze}
Zhang, X.; Park, S.; Beeler, T.; Bradley, D.; Tang, S.; and Hilliges, O. 2020.
\newblock ETH-XGaze: A large scale dataset for gaze estimation under extreme
  head pose and gaze variation.
\newblock In \emph{Proceedings of the European Conference on Computer Vision
  (ECCV)}, 365--381. Springer.

\bibitem[{Zhang et~al.(2015)Zhang, Sugano, Fritz, and Bulling}]{gazenet}
Zhang, X.; Sugano, Y.; Fritz, M.; and Bulling, A. 2015.
\newblock Appearance-based gaze estimation in the wild.
\newblock In \emph{Proceedings of the IEEE conference on computer vision and
  pattern recognition}, 4511--4520.

\bibitem[{Zhang et~al.(2017)Zhang, Sugano, Fritz, and Bulling}]{mpii}
Zhang, X.; Sugano, Y.; Fritz, M.; and Bulling, A. 2017.
\newblock It’s Written All Over Your Face: Full-Face Appearance-Based Gaze
  Estimation.
\newblock \emph{2017 IEEE Conference on Computer Vision and Pattern Recognition
  Workshops (CVPRW)}, 2299--2308.

\end{thebibliography}

\end{document}